\documentclass[]{ceurart}
\usepackage{subcaption}
\usepackage{listings}
\usepackage{graphicx} 
\usepackage[dvipsnames]{xcolor}
\usepackage{array,booktabs,arydshln}
\usepackage{amsmath, amssymb}
\usepackage{float}
\usepackage{enumitem}
\usepackage{microtype}
\usepackage{inconsolata}
\usepackage{multirow}
\usepackage{rotating}
\usepackage{tikz}
\usepackage{pifont}
\newcolumntype{C}[1]{>{\centering\arraybackslash}p{#1}}
\makeatletter
\newcommand\notsotiny{\@setfontsize\notsotiny{6.5}{7.5}}

\makeatother
\begin{document}

\copyrightyear{2023}
\copyrightclause{Copyright for this paper by its authors.Use permitted under Creative Commons License Attribution 4.0 International (CC BY 4.0).}

\conference{KBC-LM’23: Knowledge Base Construction from Pre-trained Language Models workshop at ISWC 2023}

\title{Language Models as Knowledge Bases for Visual Word Sense Disambiguation}

\author[1]{Anastasia Kritharoula}[%
orcid=0000-0002-0877-7063,
email=anaskrith@gmail.com,
]
\cormark[1]
\address[1]{Artificial Intelligence and Learning Systems Laboratory,
School of Electrical and Computer Engineering, National Technical University of Athens}

\author[1]{Maria Lymperaiou}[%
orcid=0000-0001-9442-4186,
email=marialymp@islab.ntua.gr,
]
\cormark[1]

\author[1]{Giorgos Stamou}[orcid= 0000-0003-1210-9874, email=gstam@cs.ntua.gr]

\cortext[1]{Corresponding authors.}

\begin{abstract}
Visual Word Sense Disambiguation (VWSD) is a novel challenging task that lies between linguistic sense disambiguation and fine-grained multimodal retrieval. The recent advancements in the development of visiolinguistic (VL) transformers suggest some off-the-self implementations with encouraging results, which however we argue that can be further improved. To this end, we propose some knowledge-enhancement techniques towards improving the retrieval performance of VL transformers via the usage of Large Language Models (LLMs) as Knowledge Bases. More specifically, knowledge stored in LLMs is retrieved with the help of appropriate prompts in a zero-shot manner, achieving performance advancements. Moreover, we convert VWSD to a purely textual question-answering (QA) problem by considering generated image captions as multiple-choice candidate answers. Zero-shot and few-shot prompting strategies are leveraged to explore the potential of such a transformation, while Chain-of-Thought (CoT) prompting in the zero-shot setting is able to reveal the internal reasoning steps an LLM follows to select the appropriate candidate. In total, our presented approach is the first one to analyze the merits of exploiting knowledge stored in LLMs in different ways to solve WVSD.
\end{abstract}


\maketitle

\section{Introduction}
Visual Word Sense Disambiguation (VWSD) is a newly introduced task, where the most appropriate candidate image has to be retrieved given an ambiguous target word within a specific context \cite{vwsd}. For example, given the ambiguous target word "andromeda" 
within the context "tree", the phrase "andromeda tree" is formulated. The word "andromeda" itself may have different meanings, referring to a constellation, flower, reptile or fish, and this ambiguity is mitigated via the given context "tree". However, the presence of other tree-related  candidate images complicates the choice of the golden "andromeda tree" image.

VWSD can be viewed as a multimodal extension of textual word sense disambiguation, which has received some considerable attention in recent years \cite{wsd-survey}. Resolving polysemy has been approached in several ways, with one of the most prominent being knowledge-based disambiguation. According to it, knowledge graphs representing linguistic relationships such as WordNet \cite{wordnet} for English and BabelNet \cite{babelnet} for multilingual disambiguation are exploited to frame the meaning of an ambiguous word based on its context, alleviating the need for training during the disambiguation process. More specifically, both WordNet and BabelNet knowledge graphs utilize synsets, i.e. sets of synonyms which express different meanings within different contexts; hence, this structure is inherently related to the problem of sense disambiguation, since several possible meanings are already explicitly encoded within each knowledge graph.

Nevertheless, given the ever-increasing amounts of linguistic information that need to be stored in a knowledge base, structures such as WordNet and BabelNet naturally face scalability limitations: novel words, contexts or languages demand high-quality manual engineering of the related knowledge graphs to accommodate tasks such as word sense disambiguation in a viable way. For example, given the example phrase "andromeda tree", a knowledge base lacking the target word "andromeda" or the context "tree" (with respect to "andromeda") would inevitably lead to disambiguation failure. 
To this end, the introduction of the \textit{Language Models as Knowledge Bases (LM-as-KB)} paradigm \cite{petroni2019language} opened a whole new world of possibilities for knowledge representation. The expressive power of language models allows the extraction of rich relational knowledge, similar to knowledge obtained when querying a structured knowledge graph. Despite recent research advancements, the real-world utilization of the LM-as-KB paradigm is still in its infancy, with several challenges and limitations arising \cite{lm-as-kb}. These challenges interrogate the adoption of the LM-as-KB framework in the visiolinguistic (VL) setting \cite{lymperaiou2023contribution}, even though it is already favored in some  tasks, such as Visual Question Answering \cite{vqa-lm1, vqa-lm2, vqa-lm3},  Visual Commonsense Reasoning \cite{vcr10, vcr11} and Image Captioning \cite{clipcap}.

In this work, we aim to explore the potential of following the LM-as-KB paradigm towards resolving VWSD challenges. Specifically, we focus our contributions on two main directions:

\begin{itemize}
    \item Ambiguous phrases are enriched with knowledge stored in Large Language Models (LLMs) in a zero-shot fashion to provide more context for multimodal retrieval.
    \item We explore the conversion of VWSD to textual question-answering (QA) upon which we apply zero-shot and few-shot prompting strategies. We further investigate the usage of Chain-of-Thought (CoT) prompting to obtain explanations for LLM-based decisions.
\end{itemize}

\section{Related work}
\paragraph{Language models as Knowledge Bases} is a novel paradigm that harnesses the implicit knowledge stored in neural weights of Language Models (LMs) similar to how explicit knowledge of Knowledge Graphs (KGs) has served related applications \cite{petroni2019language, lm-as-kb}. Recent works probe characteristics of different LLM-stored knowledge senses, such as commonsense \cite{commonsense-lms}, factual \cite{factual-lms} and temporal \cite{temporal-lms}.
Model scale has revealed unprecedented LM capabilities related to knowledge, such as different types of reasoning; there are open research questions related to arithmetic \cite{mathqa}, symbolic \cite{cot-reasoning}, commonsense \cite{commonsense-survey} and causal reasoning \cite{causal-reasoning} in large LMs, and whether such models merely overfit large amounts of data or are genuinely capable of reasoning similar to humans \cite{reasoning}.
More recently, the LM-as-KB paradigm has been favored by the VL community to enhance popular VL tasks \cite{lymperaiou2023contribution, vqa-lm1, vqa-lm2, vqa-lm3, vcr10, vcr11}, as in the case of the current paper.
\paragraph{Prompting Language Models}
Traditionally, LMs are pre-trained on large amounts of text corpora and then fine-tuned on smaller labelled ones to address longstanding challenges in NLP. Scaling-up to several billion parameters (Large Language Models - LLMs) enables emergent model capabilities, which can be communicated via \textit{prompting} \cite{pretrain-prompt-predict-survey}. Different prompting strategies have been proposed, such as zero-shot prompting, where the task is explicitly described in natural language, and few-shot prompting, where few exemplars are provided to the LM to guide answer generation \cite{few-shot}.
More specifically, in-context learning (ICL) is a novel few-shot paradigm where few sample demonstrations from the dataset are used to retrieve related knowledge stored in the LLM without updating any parameters \cite{in-context-survey}. This technique is inspired from the way humans learn from analogy \cite{analogy} and has been successfully applied ever since to serve several NLP tasks \cite{shi2022language, bigbench, sia2023incontext, sql, leasttomost}. Another human-inspired prompting technique is Chain-of-Thought (CoT), where the LM is encouraged by the prompt phrasing to generate rationales together with the predicted answer \cite{cot, selfconsistency, step-by-step}. Apart from the aforementioned \textit{discrete} prompt strategies, \textit{soft prompting} sacrifices human-interpretable templates to achieve more advanced knowledge retrieval \cite{soft}. In total, prompting can be viewed to be analogous to querying knowledge graphs in the traditional knowledge retrieval scenario.

\section{Method}
The retrieval backbones utilized in all cases are pre-trained VL transformers, which place both images $i$ and textual phrases $t$ in a joint embedding space. The retrieval of the most appropriate image $i$ with respect to a given text phrase $t$ is performed based on a similarity score $\textit{score}(t, i) = \textit{max}(\textit{sim}(t, i))$, which can be obtained using different similarity measures, such as cosine similarity or euclidean/manhattan distance\footnote{We will be using the term \textit{similarity} to refer to both cosine similarity and euclidean/manhattan distance.}.
In the baseline case, no knowledge enhancement is performed, thus image candidates $i$ and original phrases $t$ are passed in the VL transformer of choice. The VL transformers used are CLIP-L (CLIP \cite{clip} with ViT \cite{vit} large encoder), 
CLIP$_{LAION}$ (LAION/CLIP ViT-H/14)\cite{clip-laion} trained on LAION-2B English subset of LAION-5B,
ALIGN \cite{align}, and BLIP \cite{BLIP} variants; specifically for BLIP, BLIP$_C$ and BLIP-L$_C$ refer to variants pre-trained on COCO \cite{coco} and using ViT base/ViT large as backbone encoders respectively, while BLIP$_F$ and BLIP-L$_F$ variants are pre-trained on Flickr30k \cite{flickr}.

Moreover, based on the observations of \citet{semeval}, there are image candidates $i$ that present high similarity scores with multiple text phrases $t$, thus imposing a majority bias. To this end, we adopt the penalty factor $p(i)$, as introduced in \cite{semeval}, to diminish the similarity scores for such frequently similar candidates $i$, without affecting the similarity scores for the rest of the candidates. In cases $p(i)$ is incorporated in VL retrieval, the similarity score is altered as:
\begin{equation}\label{eq:penalty}
    score(t, i) = \textit{sim}(t, i) - p(i)
\end{equation}

\subsection{LLM-based phrase enhancement}
\label{sec:method-enhancement}
Following the LM-as-KB paradigm, the original text phrases $t$ are transformed to knowledge-enhanced phrases $t_e$ by harnessing the rich factual knowledge stored in state-of-the-art LLMs. Zero-shot enhancement is achieved via the usage of discrete prompts that follow specific handcrafted templates, as presented in Tab. \ref{tab:LLM-templates}. Those templates are created based on human conversational intuition, encouraging the retrieval of related facts to produce $t_e$. 

\begin{table*}[h!]
  \caption{Prompt templates used for zero-shot LLM knowledge enhancement of textual phrases $t$.}
  \label{tab:LLM-templates}
  \begin{tabular}{c|c}
\hline
\footnotesize \textbf{Prompt name} & \footnotesize \textbf{Prompt template} \\
\hline
\footnotesize exact &  \footnotesize “<phrase> ” \\
\hline
\footnotesize what\_is & \footnotesize “What is <phrase>?” \\
\hline
\footnotesize describe & \footnotesize “Describe <phrase>.” \\
\hline
\footnotesize meaning\_of & \footnotesize “What is the meaning of <phrase>?” \\
 \hline
\end{tabular}
\end{table*}

Emerging capabilities of LLMs, such as multiple types of reasoning, are 
analogous to model size \cite{step-by-step, wei2023chainofthought}, demonstrating that models beyond a specific scale may contain more advanced knowledge. We examine whether this statement also stands for knowledge enrichment by experimenting with models up to 13B parameters, which correspond to the upper limit our hardware can accommodate,
as well as with orders of magnitude larger models (175B parameters) accessible via public APIs (susceptible to pre-defined pricing schemes). More specifically, we use GPT2-XL (1.5B parameters) \cite{gpt2},
BLOOMZ-1.7B \& 3B \cite{bloomz},
OPT-2.7B \& 6.7B \cite{opt},
Galactica 6.7B \cite{GALACTICA},  LLAMA-7B \cite{tllama}
and Vicuna 7B \& 13B \cite{vicuna}
in the lower-billion scale, as well as the 175B parameter models of GPT-3 \cite{gpt3} and GPT-3.5-turbo{\footnote{\href{https://platform.openai.com/docs/models/gpt-3-5}{https://platform.openai.com/docs/models/gpt-3-5}}}. 

As in the baseline case, we attempt to incorporate the penalty $p(i)$ in the VL retrieval stage, forming the following knowledge-enhanced similarity score for images $i$ and phrases $t_e$:
\begin{equation}\label{eq:penalty-LLM}
    \textit{score}(t_e, i) = \textit{sim}(t_e, i) - p(i)
\end{equation}

\subsection{Prompting LLMs with question-answering prompts} 
We transform VWSD to a question-answering (QA) task by converting the textual phrases $t$ to questions $Q$ that follow handcrafted prompt templates, as presented in Tab. \ref{tab:cot}. 
Our experimentation includes both \textit{zero-shot} and \textit{few-shot} prompting. In both scenarios, the LLMs prompted are the Vicuna-13B \cite{vicuna} and the 175B GPT-3.5-turbo.
Since the LLMs to be prompted can currently handle textual but not VL inputs, we need to transit to exclusively textual representations for both images $i$ and phrases $t$. Therefore, image captioning techniques are leveraged to achieve this transformation, providing captions $c_i$ for each image candidate $i$. We experiment with greedy search, where only one caption $c_i$ per $i$ is returned, as well as with beam-search multinomial sampling using 5 beams, which returns $k$=10 captions $c_i^k$ per image $i$. The models selected for captioning are GiT-L \cite{GiT} and BLIP-L \cite{BLIP}, which are both based on ViT-large \cite{vit} encoder, and ViT-GPT2 \cite{vit-gpt}, which uses ViT-base as the encoder and GPT-2 \cite{gpt2} as the decoder. Throughout this VWSD as QA conversion we aim to leverage different aspects of knowledge stored in the selected LLMs by employing the expressiveness of the designed QA prompts, which allow for advanced flexibility compared to the  knowledge enhancement setting described in Sec. \ref{sec:method-enhancement}.

Reasoning capabilities of LLMs can be unlocked via the so-called Chain-of-Thought (CoT) prompting \cite{step-by-step, selfconsistency, cot}, where the LLM is asked to output a series of intermediate reasoning steps that logically lead to its answer. Even though CoT prompting has mostly been exploited for multi-step reasoning tasks, it is also able to provide human-understandable \textit{explanations} regarding the choice of the most appropriate candidate $i$ for each phrase $t$. To this end, the first 5 templates and the prompting pipelines of Tab. \ref{tab:cot} are adopted from \citet{cot}, where a "reasoning" prompt (\textit{"Let's think step by step"}/\textit{think} prompts of Tab. \ref{tab:cot}) retrieves the reasoning path stored in the LLM, followed by an "answer" prompt (\textit{"Therefore, [...] the answer is"}/\textit{CoT} prompt of Tab. \ref{tab:cot}) that returns the final answer in an appropriate format. The rest of the templates ("choose" prompt names) are inspired from LangChain prompts \cite{langchain}, following an instructive and descriptive template to encourage the selection of the correct image candidate.

\begin{table}[t!]
\vspace{-10px}
  \caption{QA prompts with and without CoT. 'Beam' and 'Greedy' refer to the corresponding captioning strategy.}
  \label{tab:cot}
\begin{tabular}{p{1cm}|p{13.2cm}}
\hline
\hspace{-5px}\footnotesize \textbf{Name} & \hspace{5cm} \footnotesize \textbf{QA Prompt template} \\
\hline
\hspace{-5px}\footnotesize think \newline \hspace{-5px}(greedy) &  \footnotesize “Q: What is the most appropriate caption for the <context>?
Answer choices: (A) <caption for image 1> (B) <caption for image 2> ...
A: Let’s think step by step. ”
\\
\hline
\hspace{-5px}\footnotesize think \newline \hspace{-5px}(beam)& \footnotesize “Q: What is the most appropriate group of captions for the <context>?
Answer choices: (A) <captions for image 1 (separated with comma)> (B) <captions for image 2> ...
A: Let’s think step by step. ”
\\
\hline
\hspace{-5px}\footnotesize CoT & \footnotesize “<think\_prompt> <response of LLM with think prompt>
Therefore, among A through J, the answer is”
\\
\hline
\hspace{-5px}\footnotesize no\_CoT \newline \hspace{-5px}(greedy) & \footnotesize “Q: What is the most appropriate caption for the <context>?
Answer choices: (A) <caption for image 1> (B) <caption for image 2> ...
A: ”
\\
\hline
\hspace{-5px}\footnotesize no\_CoT \newline \hspace{-5px}(beam) & \footnotesize “Q: What is the most appropriate group of captions for the <context>?
Answer choices: (A) <captions for image 1> (B) <captions for image 2> ...
A: ”
\\
\hline 
\hspace{-5px}\footnotesize choose \newline \hspace{-5px}no\_CoT \newline\hspace{-5px} \footnotesize (greedy) & \footnotesize You have ten images, (A) to (J), which are given to you in the form of captions.(A) <caption for image 1>…(J) <caption for image 10>
You should choose the image, and therefore the caption that could better represent the <phrase>.
What image do you choose?
\\
\hline 
\hspace{-5px}\footnotesize choose \newline\hspace{-5px}no\_CoT \newline\hspace{-5px} \footnotesize (beam) & \footnotesize You have ten images, (A) to (J), which are given to you in the form of captions.(A) <captions for image 1 (separated with comma)>…(J) <captions for image 10 (separated with comma)>
You should choose the image, and therefore the set of captions that could better represent the <phrase>.
What image do you choose?
\\
\hline 
\hspace{-5px}\footnotesize choose CoT (greedy) & \footnotesize You have ten images, (A) to (J), which are given to you in the form of captions.
(A) <caption for image 1>
…
(J) <caption for image 10>
You should choose the image, and therefore the caption that could better represent the <phrase>.

Use the following format:
Question: What image do you choose?
Thought: you should always think about what you choose.
Result: the result of your thought.
Final Answer: the image that you choose.

Begin!
Question: What image do you choose? 
\\
\hline 
\hspace{-6px}\footnotesize choose CoT (beam) & \footnotesize You have ten images, (A) to (J), which are given to you in the form of a set of captions.
(A) captions for image 1 (separated with comma)
…
(J) captions for image 10 (separated with comma)
You should choose the image, and therefore the set of captions that could better represent the <phrase>.

Use the following format:
Question: What image do you choose?
Thought: you should always think about what you choose.
Result: the result of your thought.
Final Answer: the image that you choose

Begin!
Question: What image do you choose? 
\\
\hline
\end{tabular}
\end{table}

\subsubsection{Zero-shot prompting}
In the \textit{zero-shot} setting, we input a selected prompt from Tab. \ref{tab:cot} to the LLM (Vicuna-13B/GPT-3.5-turbo), which generates the answer \textbf{A}. The generated \textbf{A} can be one of the caption options A-J or a statement that an answer cannot be defined; in any case, \textbf{A} is compared with the ground truth caption to determine the success or failure of the zero-shot prompting strategy under investigation. CoT prompts are accompanied with producing an explanation for choosing \textbf{A}.

\subsubsection{Few-shot prompting}
\label{sec:method-few-shot}
Additionally, we experiment with \textit{few-shot prompting} in place of the previously described \textit{zero-shot prompting}. In this case, we select \textit{k} \textit{no\_CoT/choose no\_CoT (greedy)} prompts (Tab. \ref{tab:cot}), accompanied by their ground truth \textbf{A}, thus forming \textit{QA in-context samples}. The number of in-context samples \textit{k} is defined by the user. We design three different ways of selecting the \textit{k} in-context samples. In the \textit{baseline} case (\textit{random}), the \textit{k} samples are randomly selected from the dataset. Nevertheless, since the relevance of selected samples with respect to a chosen sample is significant \cite{icl}, as well as the sample ordering \cite{prompt-ordering}, we design two similarity-based sample selection methods, namely \textit{top} and \textit{inverse-top}. Both selection methods exploit embedding representations of full phrases \textit{t}, which are obtained using ALIGN \cite{align}. The \textit{k} nearest embeddings to a given phrase embedding are efficiently retrieved with the help of cosine similarity. Then, the \textit{top} ordering strategy places the top-1 most similar QA sample first (which corresponds to the most similar full phrase embedding), followed by the 2nd most similar QA sample, until the k-th most similar in the k-th position. On the other hand, the \textit{inverse-top} strategy reverses this order by placing the top-1 most similar QA example in the k-th position.

\section{Experiments}
\subsection{LLM-based phrase enhancement results}
\label{sec:llm-enrichment}
In Tab. \ref{tab:llm-results} we present results regarding LLM-based phrase enhancement using different LLMs-as-KB and the prompts of Tab. \ref{tab:LLM-templates}. More results are provided in Appendix \ref{sec:appendix1}.

\begin{table*}[h!]
\hspace{-15px}
\vspace{-10px}
\caption{Results for zero-shot LLM-based enhancement using prompts of Tab. \ref{tab:LLM-templates}. \textcolor{RubineRed}{\textbf{Colored}} instances denote overall best results per metric, while \textbf{bold} numbers indicate best results for each LLM.}
\label{tab:llm-results}
\hskip -12cm

\begin{tabular}{p{0.0001cm}|p{1.52cm}|p{0.45cm}p{0.55cm}|p{0.45cm}p{0.55cm}|p{0.45cm}p{0.55cm}|p{0.45cm}p{0.55cm}|p{0.45cm}p{0.55cm}|p{0.45cm}p{0.55cm}|p{0.45cm}p{0.4cm}}
\hline
\multicolumn{2}{c|}{}& \multicolumn{2}{c|}{\footnotesize \textbf{CLIP$_{LAION}$}}  
& \multicolumn{2}{c|}{\footnotesize \textbf{CLIP-L}} 
& \multicolumn{2}{c|}{\footnotesize \textbf{ALIGN}} 
&  \multicolumn{2}{c|}{\footnotesize \textbf{BLIP}$_{\footnotesize \textit{C}}$} 
& \multicolumn{2}{c|}{\footnotesize \textbf{BLIP-L}$_{\footnotesize \textit{C}}$} 
& \multicolumn{2}{c|}{\footnotesize \textbf{BLIP}$_{\footnotesize \textit{F}}$} 
& \multicolumn{2}{c}{\footnotesize \textbf{BLIP-L}$_{\footnotesize \textit{F}}$}\\
\hline
\multicolumn{2}{c|}{}& \footnotesize \textbf{acc.} & \footnotesize \textbf{MRR} & \footnotesize \textbf{acc.} & \footnotesize \textbf{MRR} & \footnotesize \textbf{acc.} & \footnotesize \textbf{MRR} & \footnotesize \textbf{acc.} & \footnotesize \textbf{MRR} & \footnotesize \textbf{acc.} & \footnotesize \textbf{MRR} & \footnotesize \textbf{acc.} & \footnotesize \textbf{MRR} & \footnotesize \textbf{acc.} & \footnotesize \textbf{MRR}  \\
\hline

\multicolumn{16}{c}{\footnotesize \textbf{With penalty}} \\
\hline
\multicolumn{2}{c|}{\footnotesize \textbf{Baseline}} & \footnotesize 71.06 & \footnotesize 81.50 & \footnotesize 62.85 & \footnotesize 76.24  & \footnotesize 68.90 & \footnotesize 80.00 & \footnotesize 60.90 & \footnotesize 74.33 & \footnotesize 64.58 & \footnotesize 77.51 & \footnotesize 60.47 & \footnotesize 73.87 & \footnotesize 69.76 & \footnotesize 80.42
\\
\hline
\multirow{4}{0.08cm}{\vspace{0.2cm}\hspace{-0.12cm}\begin{turn}{90} \scriptsize \textbf{Vicuna7B} \end{turn}}& 
\footnotesize exact & \footnotesize  64.58 & \footnotesize  77.20 &
\footnotesize  59.18 & \footnotesize  72.46 &
\footnotesize  61.99 & \footnotesize  73.62 &
\footnotesize  55.94 & \footnotesize  70.98 &
\footnotesize  59.83 & \footnotesize  73.52 &
\footnotesize  53.56 & \footnotesize  69.18 &
\footnotesize  62.20 & \footnotesize  75.68 
\\
& \footnotesize what\_is & \footnotesize  69.76 & \footnotesize  80.49 &
\footnotesize  69.05 & \footnotesize  77.82 &
\footnotesize  68.25 & \footnotesize  80.16 &
\footnotesize  66.74 & \footnotesize  78.00 &
\footnotesize  69.33 & \footnotesize  80.00 &
\footnotesize  61.34 & \footnotesize  74.61 &
\footnotesize  70.84 & \footnotesize  81.00 
\\
& \footnotesize describe & \footnotesize  72.79 & \footnotesize  82.26 &
\footnotesize  68.47 & \footnotesize  78.01 &
\footnotesize  70.41 & \footnotesize  79.79 &
\footnotesize  68.25 & \footnotesize  79.09 &
\footnotesize  70.41 & \footnotesize  80.85 &
\footnotesize  62.20 & \footnotesize  75.09 &
\footnotesize  \textbf{73.43} & \footnotesize  \textbf{82.45} 
\\
& \footnotesize meaning\_of 
& \footnotesize  70.41 & \footnotesize  81.13 
& \footnotesize   67.76 & \footnotesize  77.85 &
\footnotesize  69.76 & \footnotesize  79.49 &
\footnotesize  65.87 & \footnotesize  77.84 &
\footnotesize  66.31 & \footnotesize  78.43 &
\footnotesize  62.20 & \footnotesize  74.73 &
\footnotesize  68.03 & \footnotesize  79.40 
\\
\hline
\multirow{4}{0.08cm}{\vspace{0.2cm}\hspace{-0.12cm}\begin{turn}{90} \scriptsize \textbf{Vicuna13B} \end{turn}}& 
\footnotesize exact & \footnotesize  67.60 & \footnotesize  79.48 & \footnotesize  62.12 & \footnotesize  74.49 &
\footnotesize  64.15 & \footnotesize  75.60 &
\footnotesize  60.69 & \footnotesize  73.98 &
\footnotesize  65.01 & \footnotesize  77.46 &
\footnotesize  54.86 & \footnotesize  69.91 &
\footnotesize  66.74 & \footnotesize  78.74 
\\
& \footnotesize what\_is & \footnotesize  72.14 & \footnotesize  81.69 & \footnotesize  69.98 & \footnotesize  77.63 &
\footnotesize  70.84 & \footnotesize  80.58 &
\footnotesize  67.82 & \footnotesize  79.01 &
\footnotesize  69.11 & \footnotesize  79.87 &
\footnotesize  59.18 & \footnotesize  73.48 &
\footnotesize  \textbf{73.43} & \footnotesize  \textbf{82.91} 
\\
& \footnotesize describe & \footnotesize  69.98 & \footnotesize  80.70 & \footnotesize  63.28 & \footnotesize  77.01 &
\footnotesize  66.74 & \footnotesize  76.85 &
\footnotesize  61.12 & \footnotesize  74.73 &
\footnotesize  66.95 & \footnotesize  78.57 &
\footnotesize  56.16 & \footnotesize  71.10 &
\footnotesize  67.17 & \footnotesize  79.11 
\\
& \footnotesize meaning\_of & \footnotesize  70.63 & \footnotesize  81.33 & \footnotesize  68.26 & \footnotesize  78.25 &
\footnotesize  70.63 & \footnotesize  79.83 &
\footnotesize  67.82 & \footnotesize  78.70 &
\footnotesize  69.76 & \footnotesize  80.66 &
\footnotesize  61.99 & \footnotesize  75.08 &
\footnotesize  71.71 & \footnotesize  81.80 
\\
\hline
\multirow{4}{0.08cm}{\vspace{0.2cm}\hspace{-0.15cm}\begin{turn}{90}
\scriptsize \textbf{GPT-3.5} \end{turn}}&
\footnotesize exact & \footnotesize  64.36 & \footnotesize  75.38 &
\footnotesize  60.18 & \footnotesize  72.73 &
\footnotesize  62.42 & \footnotesize  74.43 &
\footnotesize  57.02 & \footnotesize  70.78 &
\footnotesize  59.18 & \footnotesize  72.32 &
\footnotesize  52.92 & \footnotesize  67.40 &
\footnotesize  63.07 & \footnotesize  74.65 
\\
& \footnotesize what\_is & \footnotesize  70.63 & \footnotesize  81.46 &
\footnotesize  69.35 & \footnotesize  80.51 &
\footnotesize  70.41 & \footnotesize  81.42 &
\footnotesize  67.60 & \footnotesize  78.56 &
\footnotesize  68.47 & \footnotesize  79.67 &
\footnotesize  60.91 & \footnotesize  74.30 &
\footnotesize  71.71 & \footnotesize  82.02
\\
& \footnotesize describe & \footnotesize  73.22 & \footnotesize  82.50 &
\footnotesize  69.28 & \footnotesize  80.31 &
\footnotesize  73.22 & \footnotesize  82.73 &
\footnotesize  69.33 & \footnotesize  79.90 &
\footnotesize  70.41 & \footnotesize  80.80 &
\footnotesize  59.83 & \footnotesize  73.65 &
\footnotesize  70.63 & \footnotesize  81.29 
\\
& \footnotesize meaning\_of & \footnotesize  \textcolor{RubineRed}{\textbf{73.65}} & \footnotesize  \textbf{82.71} &
\footnotesize  69.06 & \footnotesize  80.55 &
\footnotesize  70.41 & \footnotesize  81.38 &
\footnotesize  66.52 & \footnotesize  78.59 &
\footnotesize  66.52 & \footnotesize  79.16 &
\footnotesize  58.53 & \footnotesize  73.31 &
\footnotesize  69.98 & \footnotesize  81.46 
\\
\hline
\multirow{4}{0.08cm}{\vspace{-0.13cm}\hspace{-0.15cm}\begin{turn}{90}
\scriptsize \textbf{GPT-3} \end{turn}}& \footnotesize exact & \footnotesize  68.03 & \footnotesize  78.41 &
\footnotesize  64.07 & \footnotesize  76.58 &
\footnotesize  66.52 & \footnotesize  78.37 &
\footnotesize  60.48 & \footnotesize  73.99 &
\footnotesize  64.15 & \footnotesize  76.58 &
\footnotesize  59.61 & \footnotesize  72.91 &
\footnotesize  65.23 & \footnotesize  77.06 
\\
&\footnotesize what\_is &\footnotesize  72.35 & \footnotesize 82.19 &
\footnotesize  70.73 & \footnotesize  81.57 &
\footnotesize  71.71 & \footnotesize  82.27 &
\footnotesize  68.25 & \footnotesize  78.93 &
\footnotesize  68.90 & \footnotesize  79.91 &
\footnotesize  60.48 & \footnotesize  74.24 &
\footnotesize  69.11 & \footnotesize  80.25 
\\
& \footnotesize describe &\footnotesize  70.63 & \footnotesize  81.05 &
\footnotesize  68.72 & \footnotesize  80.26 &
\footnotesize  72.57 & \footnotesize  82.52 &
\footnotesize  64.58 & \footnotesize  76.75 &
\footnotesize  68.25 & \footnotesize  79.35 &
\footnotesize  61.34 & \footnotesize  74.03 &
\footnotesize  69.33 & \footnotesize  80.47 
\\
& \footnotesize meaning\_of &\footnotesize  \textcolor{RubineRed}{\textbf{73.65}} & \footnotesize  \textcolor{RubineRed}{\textbf{83.52}} &
\footnotesize  69.84 & \footnotesize  81.56 &
\footnotesize  74.95 & \footnotesize  84.09 &
\footnotesize  66.74 & \footnotesize  78.37 &
\footnotesize  71.71 & \footnotesize  81.55 &
\footnotesize  62.63 & \footnotesize  75.55 &
\footnotesize  72.35 & \footnotesize  82.28 
\\
\hline

\hline
\multicolumn{16}{c}{\footnotesize \textbf{Without penalty}} \\
\hline
\multicolumn{2}{c|}{\footnotesize \textbf{Baseline}} & \footnotesize 67.82 & \footnotesize 79.50 & \footnotesize 60.69 & \footnotesize 74.42 & \footnotesize 65.66 & \footnotesize 77.48 & \footnotesize 57.24 & \footnotesize 72.07 & \footnotesize 61.34 & \footnotesize 75.88 & \footnotesize 57.67 & \footnotesize 71.96 & \footnotesize 65.01 & \footnotesize 77.86
\\
\hline
\multirow{4}{0.08cm}{\vspace{0.2cm}\hspace{-0.12cm}\begin{turn}{90} \scriptsize \textbf{Vicuna7B} \end{turn}}& 
\footnotesize exact & \footnotesize  61.99 & \footnotesize  75.30 &\footnotesize  57.45 & \footnotesize  70.07 &
\footnotesize  58.96 & \footnotesize  71.92 &
\footnotesize  52.70 & \footnotesize  68.00 &
\footnotesize  56.16 & \footnotesize  71.28 &
\footnotesize  50.32 & \footnotesize  66.75 &
\footnotesize  60.04 & \footnotesize  74.04 
\\
& \footnotesize what\_is & 
\footnotesize  68.47 & \footnotesize  79.28 &
\footnotesize  65.58 & \footnotesize  76.70 &
\footnotesize  64.58 & \footnotesize  77.93 &
\footnotesize  64.79 & \footnotesize  76.50 &
\footnotesize  66.52 & \footnotesize  78.11 &
\footnotesize  57.88 & \footnotesize  72.77 &
\footnotesize  67.17 & \footnotesize  78.98 
\\
& \footnotesize describe & \footnotesize  \textbf{71.06} & \footnotesize  \textbf{81.17} &
\footnotesize  65.87 & \footnotesize  76.16 &
\footnotesize  68.25 & \footnotesize  77.91 &
\footnotesize  66.09 & \footnotesize  77.78 &
\footnotesize  68.90 & \footnotesize  79.63 &
\footnotesize  60.69 & \footnotesize  73.97 &
\footnotesize  70.19 & \footnotesize  80.69 
\\
& \footnotesize meaning\_of & 
\footnotesize  69.11 & \footnotesize  80.13 
&\footnotesize  65.80 & \footnotesize  75.77 &
\footnotesize  68.25 & \footnotesize  78.11 &
\footnotesize  63.28 & \footnotesize  75.79 &
\footnotesize  63.07 & \footnotesize  75.96 &
\footnotesize  59.18 & \footnotesize  72.88 &
\footnotesize  64.36 & \footnotesize  77.09 
\\
\hline
\multirow{4}{0.08cm}{\vspace{0.2cm}\hspace{-0.12cm}\begin{turn}{90} \scriptsize \textbf{Vicuna13B} \end{turn}}& 
\footnotesize exact &  \footnotesize  65.23 & \footnotesize  77.73 &
\footnotesize  58.44 & \footnotesize  71.67 &
\footnotesize  61.56 & \footnotesize  73.10 &
\footnotesize  56.80 & \footnotesize  71.48 &
\footnotesize  60.69 & \footnotesize  74.43 &
\footnotesize  51.19 & \footnotesize  66.95 &
\footnotesize  63.28 & \footnotesize  76.68  \\
& \footnotesize what\_is & \footnotesize  \textbf{70.63} & \footnotesize  \textbf{80.71} & \footnotesize  66.74 & \footnotesize  75.51 &
\footnotesize  68.68 & \footnotesize  78.55 &
\footnotesize  65.44 & \footnotesize  77.22 &
\footnotesize  68.47 & \footnotesize  79.14 &
\footnotesize  58.96 & \footnotesize  73.22 &
\footnotesize  68.90 & \footnotesize  80.27 
\\
& \footnotesize describe & \footnotesize  67.82 & \footnotesize  79.14 &\footnotesize  60.69 & \footnotesize  75.75 &
\footnotesize  64.58 & \footnotesize  74.88 &
\footnotesize  57.67 & \footnotesize  72.47 &
\footnotesize  62.20 & \footnotesize  75.75 &
\footnotesize  52.27 & \footnotesize  68.72 &
\footnotesize  63.50 & \footnotesize  76.66 
\\
& \footnotesize meaning\_of & \footnotesize  68.90 & \footnotesize  80.18 &\footnotesize  65.87 & \footnotesize  76.64 &
\footnotesize  67.60 & \footnotesize  78.38 &
\footnotesize  65.66 & \footnotesize  77.62 &
\footnotesize  66.31 & \footnotesize  78.31 &
\footnotesize  58.96 & \footnotesize  73.19 &
\footnotesize  68.03 & \footnotesize  79.64 
\\
\hline
\multirow{4}{0.08cm}{\vspace{0.2cm}\hspace{-0.15cm}\begin{turn}{90} \scriptsize \textbf{GPT-3.5} \end{turn}}& 
\footnotesize exact & \footnotesize  62.20 & \footnotesize  73.38 &
\footnotesize  57.11 & \footnotesize  70.36 &
\footnotesize  60.48 & \footnotesize  72.15 &
\footnotesize  54.43 & \footnotesize  68.33 &
\footnotesize  56.80 & \footnotesize  70.42 &
\footnotesize  51.40 & \footnotesize  65.68 &
\footnotesize  58.32 & \footnotesize  71.11 
\\
& \footnotesize what\_is & \footnotesize  69.55 & \footnotesize  80.51 &
\footnotesize  65.87 & \footnotesize  78.11 &
\footnotesize  67.82 & \footnotesize  79.52 &
\footnotesize  64.15 & \footnotesize  75.91 &
\footnotesize  65.87 & \footnotesize  77.78 &
\footnotesize  58.10 & \footnotesize  72.32 &
\footnotesize  68.03 & \footnotesize  79.36 
\\
& \footnotesize describe & \footnotesize  72.57 & \footnotesize  81.78 &
\footnotesize  66.67 & \footnotesize  78.42 &
\footnotesize  70.84 & \footnotesize  81.16 &
\footnotesize  65.44 & \footnotesize  77.57 &
\footnotesize  69.11 & \footnotesize  80.20 &
\footnotesize  58.96 & \footnotesize  72.66 &
\footnotesize  67.60 & \footnotesize  79.47 
\\
& \footnotesize meaning\_of  & \footnotesize  \textbf{72.57} & \footnotesize \textbf{82.05} &
\footnotesize  67.10 & \footnotesize  79.07 &
\footnotesize  68.47 & \footnotesize  79.87 &
\footnotesize  63.93 & \footnotesize  77.05 &
\footnotesize  65.66 & \footnotesize  78.33 &
\footnotesize  63.93 & \footnotesize  72.23 &
\footnotesize  68.25 & \footnotesize  80.17 
\\
\hline

\multirow{4}{0.08cm}{\vspace{-0.13cm}\hspace{-0.15cm}\begin{turn}{90}
\scriptsize \textbf{GPT-3} \end{turn}}& \footnotesize exact &
\footnotesize  66.74 & \footnotesize  77.22 &
\footnotesize  61.68 & \footnotesize  74.91 &
\footnotesize  64.79 & \footnotesize  76.27 &
\footnotesize  58.96 & \footnotesize  71.92 &
\footnotesize  60.48 & \footnotesize  74.02 &
\footnotesize  55.72 & \footnotesize  70.34 &
\footnotesize  62.42 & \footnotesize  75.04 
\\
&\footnotesize what\_is &\footnotesize  71.06 & \footnotesize  81.12 &
\footnotesize  68.15 & \footnotesize  79.38 &
\footnotesize  69.55 & \footnotesize  80.22 &
\footnotesize  63.28 & \footnotesize  75.56 &
\footnotesize  65.01 & \footnotesize  77.40 &
\footnotesize  56.59 & \footnotesize  71.54 &
\footnotesize  67.82 & \footnotesize  79.03 
\\
& \footnotesize describe &\footnotesize  69.55 & \footnotesize  80.15 &
\footnotesize  68.25 & \footnotesize  79.81 &
\footnotesize  71.27 & \footnotesize  81.21 &
\footnotesize  63.93 & \footnotesize  75.81 &
\footnotesize  66.31 & \footnotesize  77.74 &
\footnotesize  58.96 & \footnotesize  72.62 &
\footnotesize  67.17 & \footnotesize  78.93 
\\
& \footnotesize meaning\_of & \footnotesize  \textbf{73.00} & \footnotesize  \textbf{82.89} &
\footnotesize  68.96 & \footnotesize  80.26 &
\footnotesize  72.57 & \footnotesize 82.29 &
\footnotesize  65.87 & \footnotesize  77.56 &
\footnotesize  69.55 & \footnotesize  80.26 &
\footnotesize  60.26 & \footnotesize  74.26 &
\footnotesize  70.41 & \footnotesize  81.09 
\\
\hline
\end{tabular}

\end{table*}

We can easily observe that LLM-based enhancement helps trespassing baseline accuracy and MRR scores when appropriate prompting is used,  irrespectively of the usage of VL penalty $p(i)$; "exact" prompt seems to be rather weak towards triggering the necessary knowledge to further drive VL retrieval, as in several cases metrics corresponding to phrases enhanced using the "exact" prompt fall below the baseline performance. On the other hand, "meaning\_of" is the more powerful prompt attempted, resulting in non-negligible performance advancements compared to the baselines in most cases.
Overall, the combination of GPT-3 phrase enrichment together with CLIP$_{LAION}$ (with penalty $p(i)$) as the VL retrieval module lead to optimal results.

An interesting observation is that the Vicuna 7/13B models \textbf{perform comparably} to GPT-3/3.5 models despite being orders of magnitude smaller. This is an encouraging result suggesting that LLM-based phrase enrichment \textit{may} be successfully performed with more lightweight LLMs that do not adhere to a limiting pricing plan, which would impede large-scale experimentation. However, in most cases that LLMs in the lower-billion scale are employed for knowledge enhancement (Tab. \ref{tab:llm-results-appendix-1}), retrieval results struggle to compete with the knowledge-free baselines, revealing a non-negligible association between scale and knowledge-enhancement capabilities.
\subsection{Question-answering prompting results}

In Tab. \ref{tab:cot-results} we present accuracy scores occurring from transforming VWSD to QA using zero-shot (with and without CoT) and few-shot (without CoT) prompting, denoting best results per prompt with \textcolor{RubineRed}{\textbf{color}}.  In general, there is an obvious discrepancy between the performance of GPT-3.5-turbo and Vicuna-13B, denoting that \textbf{model scale does matter} in the VWSD as QA scenario, contrary to the phrase enhancement case (Sec. \ref{sec:llm-enrichment}): smaller models do not possess the necessary knowledge or reasoning abilities to infer the correct answer from captions in the QA setting, no matter the prompt template deployed or the choice between zero-shot/few-shot strategy. This finding agrees with the observation of \citet{cot} that LLM reasoning capabilities emerge \textit{at scale}.
Moreover, there is no clear pattern whether beam or greedy captioning is more effective towards triggering the necessary knowledge: in the case of GiT-L and BLIP-L there is a clear preference towards greedy decoding in conjunction with CoT \& no\_CoT prompting. 
Especially for Vicuna-13B this preference is very distinctive, demonstrating significant performance drops when beam decoding is employed in place of greedy decoding.
However, the opposite holds for ViT-GPT2, even though the overall performance of ViT-GPT2 is lower in comparison to the other captioners.
The performance becomes even worse when beam decoding is employed with few-shot prompting.
At the same time, "choose" prompts do not demonstrate a distinct pattern with respect to the decoding strategy (greedy/beam) as well. Overall, GiT-L (greedy) exhibits the most promising captioning capabilities in the majority of Tab. \ref{tab:cot-results} results, while BLIP-L (greedy) appears more capable in the few-shot prompting setting of Vicuna-13B (\textcolor{RubineRed}{\textbf{colored}} cells).

\begin{table}[h!]
\hspace{-10px}
\caption{Accuracy scores (\%) for VWSD as a QA problem with and without CoT prompting.}
\label{tab:cot-results}
\begin{tabular}{p{2.35cm}|>{\centering\arraybackslash}p{1cm}>{\centering\arraybackslash}p{1cm}>{\centering\arraybackslash}p{1.95cm}>{\centering\arraybackslash}p{1.47cm}|>{\centering\arraybackslash}p{1.4cm}|>{\centering\arraybackslash}p{1.33cm}|>{\centering\arraybackslash}p{1.32cm}}
\hline
\footnotesize \textbf{Captioner} & \multicolumn{4}{c|}{\footnotesize \textbf{Zero-shot}} & \footnotesize \textbf{Few-shot\newline (random)} & \footnotesize \textbf{Few-shot\newline (top)}  & \footnotesize \textbf{Few-shot\newline (inv. top)}\\
\cline{2-8}
& \footnotesize no\_CoT & \footnotesize CoT & 
\footnotesize choose no\_CoT & \footnotesize choose CoT &\footnotesize no\_CoT & \footnotesize no\_CoT  & \footnotesize no\_CoT  \\
\hline
\multicolumn{7}{c}{\footnotesize \textbf{\hspace{2cm} GPT-3.5-turbo}} \\
\hline
\footnotesize GiT-L (greedy) & \footnotesize 44.49  &	\footnotesize \textcolor{RubineRed}{\textbf{47.30}} & \footnotesize \textcolor{RubineRed}{\textbf{51.84}} & \footnotesize \textcolor{RubineRed}{\textbf{52.27}} & \footnotesize \textcolor{RubineRed}{\textbf{51.19}}  & \footnotesize \textcolor{RubineRed}{\textbf{51.40}} & \footnotesize \textcolor{RubineRed}{\textbf{53.56}}\\
\footnotesize GiT-L (beam) & \footnotesize 40.82 & \footnotesize 36.50 & \footnotesize 50.54 & \footnotesize 49.68 & \footnotesize 46.12 & \footnotesize 47.83 & \footnotesize 45.61\\
\footnotesize BLIP-L (greedy) & \footnotesize \textcolor{RubineRed}{\textbf{47.95}} 	& \footnotesize 43.84 & \footnotesize 49.46 & \footnotesize 44.06	& \footnotesize 48.16 & \footnotesize 48.81	& \footnotesize 50.32\\
\footnotesize BLIP-L (beam) &	\footnotesize 38.01  & \footnotesize 34.13 & \footnotesize 50.97 & \footnotesize 50.97  & \footnotesize 40.91 & \footnotesize 40.49 & \footnotesize 40.49\\
\footnotesize ViT-GPT2 (greedy) &	\footnotesize 28.94  & \footnotesize 25.05	& \footnotesize 32.40 & \footnotesize 29.81 & \footnotesize 31.32 & \footnotesize 31.45 & \footnotesize 28.91\\
\footnotesize ViT-GPT2 (beam) & \footnotesize 30.24 & \footnotesize 25.92 & \footnotesize 32.83 & \footnotesize 33.05& \footnotesize 32.03 & \footnotesize 28.73 & \footnotesize 23.64\\
\hline
\multicolumn{7}{c}{\footnotesize \textbf{\hspace{2cm} Vicuna-13B}} \\
\hline
\footnotesize  GiT-L (greedy) &	\footnotesize 
 \textcolor{RubineRed}{\textbf{34.34}} & \footnotesize \textcolor{RubineRed}{\textbf{27.65}} & \footnotesize 20.52 & \footnotesize 20.52 & \footnotesize  31.89 & \footnotesize  33.63 & \footnotesize  36.30\\
\footnotesize  GiT-L (beam) & \footnotesize	11.02 & \footnotesize 7.91 & \footnotesize 19.44 & \footnotesize 11.23 & \footnotesize < 2  & \footnotesize < 2 & \footnotesize < 2 \\
 \footnotesize BLIP-L (greedy) & \footnotesize 30.02 & \footnotesize 23.76 & \footnotesize \textcolor{RubineRed}{\textbf{20.95}} & \footnotesize \textcolor{RubineRed}{\textbf{21.81}} & \footnotesize  \textcolor{RubineRed}{\textbf{35.56}} & \footnotesize  \textcolor{RubineRed}{\textbf{36.08}} & \footnotesize  \textcolor{RubineRed}{\textbf{36.48}}\\
\footnotesize  BLIP-L (beam) & \footnotesize 9.41 & \footnotesize 6.27 & \footnotesize 12.74 & \footnotesize 8.64 & \footnotesize < 2  & \footnotesize < 2 & \footnotesize < 2 \\
\footnotesize ViT-GPT2 (greedy) & \footnotesize 21.60 & \footnotesize 21.17  & \footnotesize 17.49 & \footnotesize 15.33& \footnotesize 24.83 & \footnotesize 24.94 & \footnotesize 26.11\\
\footnotesize ViT-GPT2 (beam) & \footnotesize 11.45 & \footnotesize 6.91 & \footnotesize 16.85 & \footnotesize 12.74 & \footnotesize 2.81 & \footnotesize 3.89 & \footnotesize 4.75 \\
\hline
\end{tabular}
\end{table}

\subsubsection{Zero-shot prompting and CoT reasoning}
\begin{figure*}[h!]
    \centering
    \begin{subfigure}{0.19\textwidth}
        \centering
        \includegraphics[width=2.3cm, height=1.6cm]{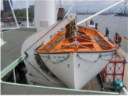}
        \caption*{\footnotesize A}
    \end{subfigure}
    \hfill
    \begin{subfigure}{0.19\textwidth}
        \centering
        \includegraphics[width=2.3cm, height=1.6cm]{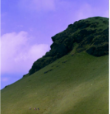}
        \caption*{\footnotesize B}
    \end{subfigure}
    \hfill
    \begin{subfigure}{0.19\textwidth}
        \centering
        \includegraphics[width=2.3cm, height=1.6cm]{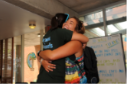}
        \caption*{\footnotesize C (Gold image)}
    \end{subfigure}
    \hfill
    \begin{subfigure}{0.19\textwidth}
        \centering
        \includegraphics[width=2.3cm, height=1.6cm]{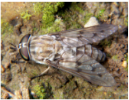}
        \caption*{\footnotesize D}
    \end{subfigure}
    \hfill
    \begin{subfigure}{0.19\textwidth}
        \centering
        \includegraphics[width=2.3cm, height=1.6cm]{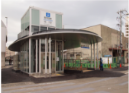}
        \caption*{\footnotesize E}
    \end{subfigure}
    
    
    \begin{subfigure}{0.19\textwidth}
        \centering
        \includegraphics[width=2.3cm, height=1.6cm]{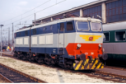}
        \caption*{\footnotesize F}
    \end{subfigure}
    \hfill
    \begin{subfigure}{0.19\textwidth}
        \centering
        \includegraphics[width=2.3cm, height=1.6cm]{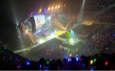}
        \caption*{\footnotesize G}
    \end{subfigure}
    \hfill
    \begin{subfigure}{0.19\textwidth}
        \centering
        \includegraphics[width=2.3cm, height=1.6cm]{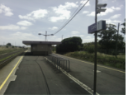}
        \caption*{\footnotesize H}
    \end{subfigure}
    \hfill
    \begin{subfigure}{0.19\textwidth}
        \centering
        \includegraphics[width=2.3cm, height=1.6cm]{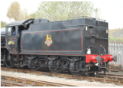}
        \caption*{\footnotesize I}
    \end{subfigure}
    \hfill
    \begin{subfigure}{0.19\textwidth}
        \centering
        \includegraphics[width=2.3cm, height=1.6cm]{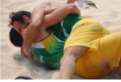}
        \caption*{\footnotesize J}
    \end{subfigure}
    
    \caption{Candidate images for the phrase \textit{t} "tender embrace".}
    \label{fig:cot-example-appendix2}
\end{figure*}

According to Tab. \ref{tab:cot-results}, accuracy results with zero-shot QA prompting are not always encouraging, especially when compared to LLM-based enrichment results of Tab. \ref{tab:llm-results}, which approach state-of-the-art performance. 
Apart from the fact that proper LLM reasoning is still an open problem \cite{bang2023multitask, turpin2023language}, one possible fundamental reason for these lower accuracy scores can be the conversion from images to text via captioning: this intra-modality conversion may induce errors and information loss that impacts the final performance. Of course, lower performance will also affect the quality of the produced explanations, when CoT is employed.
We delve into this scenario by presenting an example where the usage of CoT leads to an erroneous answer, while the no\_CoT case succeeds. 
In Fig. \ref{fig:cot-example-appendix2} the candidates for the phrase "tender embrace" are showcased. By employing GiT-L (greedy) as the image captioner, we form the question Q as:

\begin{table}[h!]
\hspace{-10px}
\vspace{-3px}
\begin{tabular}{p{14.5cm}}
\hline
\footnotesize \textit{Q: What is the most appropriate caption for the tender embrace? 
Answer Choices: (A) a small boat sitting on top of a dock. (B) a group of people walking on a green hill. (C) a student gets a hug from a student. (D) a large fly laying on a rock in the water. (E) the bus stop at the station (F) a train is parked at a station. (G) a crowd of people watching a concert. (H) a train station with a sign on the side of it. (I) a black and red train on a track. (J) a man laying in the sand on top of a surfboard.} \\
\hline
\end{tabular}
\end{table}
In Tab. \ref{tab:cot-fail-example} the results of GPT-3.5-turbo with and without \textit{zero-shot} CoT are demonstrated. 
\begin{table}[h!]
\hspace{-20cm}
\caption{Example of VWSD as QA with \& w/o zero-shot CoT prompting. Questions Q are presented truncated. GPT-3.5-turbo is the selected LLM, as it presented best results for CoT/no\_CoT prompts of Tab. \ref{tab:cot-results}.}
\label{tab:cot-fail-example}
\begin{tabular}{p{14.4cm}}
\hline
\hspace{7cm}\textbf{\footnotesize no\_CoT}    \\
\hline
\footnotesize Q: What is the most appropriate [...]

A: \textcolor{blue}{The most appropriate caption for the tender embrace would be: (C) a student gets a hug from a student.} \textcolor{ForestGreen}{\ding{52}}
\\
\hline
\hspace{7.3cm}\textbf{\footnotesize CoT} \\
\hline
\footnotesize Q: What is the most appropriate [...]

\textcolor{red}{A: Let's think step by step.}

\textcolor{blue}{First, we need to understand what the tender embrace is referring to. Once we know that, we can look at the answer choices and find the one that best matches the subject of the photo. Therefore, without further information about the photo, it is not possible to choose the most appropriate caption. Could you please provide more context or information about the photo?
}

\textcolor{red}{Therefore, among A through J, the answer is} \textcolor{blue}{not applicable without more information about the photo" }\textcolor{red}{\textbf{X}}
\\
\hline
\end{tabular}
\end{table}

After observing the captions $c_i$ produced by GiT-L, we confirm that they accurately describe the context, thus being appropriate answer options. Nevertheless, when prompted with CoT in a zero-shot manner, GPT-3.5-turbo is unable to define where "tender embrace" refers to without receiving more information. On the contrary, it successfully returns the right answer when no CoT is used. This discrepancy reveals that even though the knowledge regarding the phrase "tender embrace" exists within the LLM, zero-shot CoT prompting is unable to trigger it. This can be viewed as an inherent problem of CoT prompting, at least in the zero-shot setting, since the only factor that differs between successful and unsuccessful reasoning is the prompt itself.

\begin{figure*}[h]
    \centering
    \begin{subfigure}{0.19\textwidth}
        \centering
        \includegraphics[width=2.3cm, height=1.6cm]{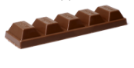}
        \caption*{\footnotesize A}
    \end{subfigure}
    \hfill
    \begin{subfigure}{0.19\textwidth}
        \centering
        \includegraphics[width=2.3cm, height=1.6cm]{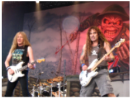}
        \caption*{\footnotesize B}
    \end{subfigure}
    \hfill
    \begin{subfigure}{0.19\textwidth}
        \centering
        \includegraphics[width=2.3cm, height=1.6cm]{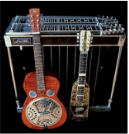}
        \caption*{\footnotesize C}
    \end{subfigure}
    \hfill
    \begin{subfigure}{0.19\textwidth}
        \centering
        \includegraphics[width=2.3cm, height=1.6cm]{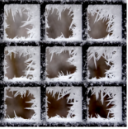}
        \caption*{\footnotesize D}
    \end{subfigure}
    \hfill
    \begin{subfigure}{0.19\textwidth}
        \centering
        \includegraphics[width=2.3cm, height=1.6cm]{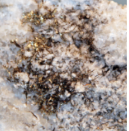}
        \caption*{\footnotesize E}
    \end{subfigure}
    
    
    \begin{subfigure}{0.19\textwidth}
        \centering
        \includegraphics[width=2.3cm, height=1.6cm]{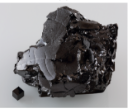}
        \caption*{\footnotesize F}
    \end{subfigure}
    \hfill
    \begin{subfigure}{0.19\textwidth}
        \centering
        \includegraphics[width=2.3cm, height=1.6cm]{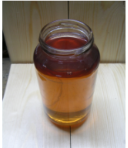}
        \caption*{\footnotesize G}
    \end{subfigure}
    \hfill
    \begin{subfigure}{0.19\textwidth}
        \centering
        \includegraphics[width=2.3cm, height=1.6cm]{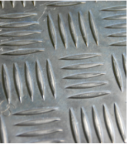}
        \caption*{\footnotesize H (Gold image)}
    \end{subfigure}
    \hfill
    \begin{subfigure}{0.19\textwidth}
        \centering
        \includegraphics[width=2.3cm, height=1.6cm]{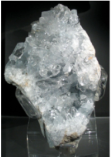}
        \caption*{\footnotesize I}
    \end{subfigure}
    \hfill
    \begin{subfigure}{0.19\textwidth}
        \centering
        \includegraphics[width=2.3cm, height=1.6cm]{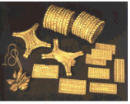}
        \caption*{\footnotesize J}
    \end{subfigure}
    
    \caption{Candidate images for the phrase \textit{t} "metal steel".}
    \label{fig:cot-example-appendix1}
\end{figure*}

On the other hand, zero-shot CoT prompting can provide valuable insights in several cases, such as the one presented in Fig. \ref{fig:cot-example-appendix1} with candidates corresponding to the phrase "metal steel".
By again using GiT-L as the captioner, the question Q is formed as:
\begin{table}[h!]
\vspace{-10px}
\begin{tabular}{p{14.5cm}}
\hline
\footnotesize \textit{Q: What is the most appropriate caption for the metal steel? 
Answer Choices: (A) a chocolate bar with three sides (B) [unused0] and [unused0] at the concert in 2007 (C) a guitar and a guitar are displayed in front of a speaker. (D) frosty patterns on a window (E) gold in the rocks - - (F) a black piece of metal with a large black square in the middle. (G) a jar of honey on a wooden table. (H) a close up of a metal plate with a pattern of lines. (I) a large white quartz rock with a clear base. (J) gold jewelry from the late 19th century.} \\
\hline
\end{tabular}
\end{table}

Then, GPT-3.5-turbo outputs the results presented in Tab. \ref{tab:cot-comparison-appendix1}, successfully guided by CoT.

\begin{table}[h!]
\caption{A successful example of VWSD as QA with \& w/o zero-shot CoT prompting (GPT-3.5-turbo as the LLM).}
\label{tab:cot-comparison-appendix1}
\begin{tabular}{p{14.4cm}}
\hline
\hspace{7cm}\textbf{\footnotesize no\_CoT}    \\
\hline
\footnotesize 
\footnotesize Q: What is the most appropriate [...]

A: \textcolor{blue}{The most appropriate caption for the metal steel would be (F) a black piece of metal with a large black square in the middle.} \textcolor{red}{\textbf{X}}
\\
\hline
\hspace{7.3cm}\textbf{\footnotesize CoT} \\
\hline
\footnotesize Q: What is the most appropriate [...]

\textcolor{red}{A: Let's think step by step.}

\textcolor{blue}{First, we need to understand what metal steel is and what its characteristics are. Steel is a hard and strong metal alloy made mainly of iron and carbon. It is often used in construction, machinery, and transportation. Based on this information, the most appropriate caption for metal steel would be (H) a close up of a metal plate with a pattern of lines. This caption describes the texture and appearance of steel, which is often characterized by its distinctive pattern of lines. The other options do not accurately describe steel or its unique qualities.
}

\textcolor{red}{Therefore, among A through J, the answer is} \textcolor{blue}{(H) a close up of a metal plate with a pattern of lines." } \textcolor{ForestGreen}{\ding{52}}
\\
\hline
\end{tabular}
\end{table}

This analysis highlights that the current state of CoT reasoning \textit{cannot be entirely trusted}, as the deployment of CoT prompts themselves may drive the LLM towards incorrect reasoning paths. Thus, CoT reasoning for VWSD is not yet mature enough to provide valuable explanations.

Furthermore, regarding the comparison between standalone CoT/no\_CoT prompts and "choose" CoT/no\_CoT prompts, there is a clear performance advancement when "choose" prompts are leveraged, reaching more than 10\% performance gain in the case of BLIP-L (beam) as captioner and GPT-3.5-turbo as the LLM to be prompted. We regard this as an evidence that more descriptive and motivating prompts can better evoke the correct reasoning process of an LLM. Finally, there is no clear indication of whether CoT facilitates performance over no\_CoT when combined with the "choose" prompting template, since best results per captioner and LLM of Tab. \ref{tab:cot-results} alternate between "choose" CoT/no\_CoT strategies, and are often comparable. 

\subsubsection{Few-shot prompting}
In the baseline few-shot setting (few-shot
(random)), we randomly select \textit{k}=5 instances to serve as in-context examples. These in-context samples are comprised of \textit{k} questions Q followed by their ground truth answer choice, as described in Sec. \ref{sec:method-few-shot}.
Results of Tab. \ref{tab:cot-results} denote that few-shot performance using no\_CoT prompts is significantly better compared to their zero-shot counterparts, while generally being on par with "choose" prompt results, even though the derived accuracy is close to random choice or even worse in most cases. Therefore, despite the advanced engineering few-shot prompting requires compared to the zero-shot setting, it can be regarded as a more viable choice for retrieving internal reasoning capabilities of LLMs. As for the choice of in-context sample selection and order strategy, the results are not conclusive; similarity-based sample selection (top \& inv. top columns of Tab. \ref{tab:cot-results}) may result in better (GIT-L greedy), worse (ViT-GPT2 beam) or comparable results to the random baseline accuracy. Moreover, sample ordering (top versus inv. top) presents some variability in accuracy, with each strategy performing better in different cases. Overall, we can state that few-shot prompting calls for more extensive experimentation beyond the purpose of the current work until a standard pattern emerges, while there is a possibility that no pattern can be inferred at all.

\begin{figure*}[h!]
    \centering
    \begin{subfigure}{0.19\textwidth}
        \centering
        \includegraphics[width=2.3cm, height=1.6cm]{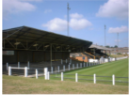}
        \caption*{\footnotesize A}
    \end{subfigure}
    \hfill
    \begin{subfigure}{0.19\textwidth}
        \centering
        \includegraphics[width=2.3cm, height=1.6cm]{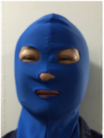}
        \caption*{\footnotesize B}
    \end{subfigure}
    \hfill
    \begin{subfigure}{0.19\textwidth}
        \centering
        \includegraphics[width=2.3cm, height=1.6cm]{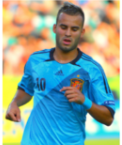}
        \caption*{\footnotesize C}
    \end{subfigure}
    \hfill
    \begin{subfigure}{0.19\textwidth}
        \centering
        \includegraphics[width=2.3cm, height=1.6cm]{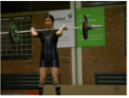}
        \caption*{\footnotesize D}
    \end{subfigure}
    \hfill
    \begin{subfigure}{0.19\textwidth}
        \centering
        \includegraphics[width=2.3cm, height=1.6cm]{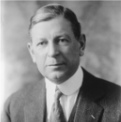}
        \caption*{\footnotesize E}
    \end{subfigure}
    
    
    \begin{subfigure}{0.19\textwidth}
        \centering
        \includegraphics[width=2.3cm, height=1.6cm]{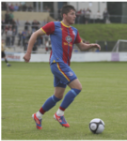}
        \caption*{\footnotesize F}
    \end{subfigure}
    \hfill
    \begin{subfigure}{0.19\textwidth}
        \centering
        \includegraphics[width=2.3cm, height=1.6cm]{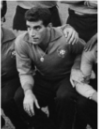}
        \caption*{\footnotesize G}
    \end{subfigure}
    \hfill
    \begin{subfigure}{0.19\textwidth}
        \centering
        \includegraphics[width=2.3cm, height=1.6cm]{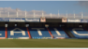}
        \caption*{\footnotesize H}
    \end{subfigure}
    \hfill
    \begin{subfigure}{0.19\textwidth}
        \centering
        \includegraphics[width=2.3cm, height=1.6cm]{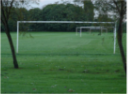}
        \caption*{\footnotesize I (Gold image)}
    \end{subfigure}
    \hfill
    \begin{subfigure}{0.19\textwidth}
        \centering
        \includegraphics[width=2.3cm, height=1.6cm]{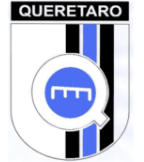}
        \caption*{\footnotesize J}
    \end{subfigure}
    
    \caption{Candidate images for the phrase \textit{t} "football goal".}
    \label{fig:few-shot}
\end{figure*}

We are going to present some qualitative results regarding few-shot prompting.
Fig. \ref{fig:few-shot} contains candidates corresponding to the phrase "football goal", while GiT-L (greedy) serves as the captioner. The \textit{k}=5 in-context samples are demonstrated in Tab. \ref{tab:in-context} followed by the answer \textbf{A} generated by GPT-3.5-turbo (in \textcolor{magenta}{color}) corresponding to the given \textbf{Q}. In the presented case, in-context prompting achieves in guiding GPT-3.5-turbo to select the correct candidate I.

\begin{table}[h!]
\caption{Example of few-shot prompting with k=5 in-context samples for disambiguating "football goal" phrase.}
\label{tab:in-context}
\begin{tabular}{p{14.7cm}}
\hline
\hspace{-10px}
\footnotesize \textbf{Q:} What is the most appropriate caption for the vanillin compound? 
Answer Choices: (A) a fossil fossil with a spiral pattern (B) a blue plate topped with pastries and a piece of bread. (C) the bones of the foot are very thin. (D) a spoon full of sugar on a table. (E) the tower of the building is made of brick. (F) a pair of sunglasses with a white background. (G) a close up of a plant with a yellow flower. (H) turmeric powder in a bowl (I) a man with glasses and a tie sitting at a desk. (J) a close up of a bike tire with a yellow tag on it. 
\textbf{A:} (D) a spoon full of sugar on a table.

\hspace{-10px}
\textbf{Q:} What is the most appropriate caption for the nankeen fabric? 
Answer Choices: (A) a close up of a white quilt with a tree design on it. (B) a man in a blue jacket and tan pants holding a sword. (C) a collection of brass trombones (D) a close up of a green plant (E) a pine tree in the forest (F) a row of mannequins in black dresses. (G) two boys are standing in front of a red wall. (H) two apples sitting next to each other on a white surface. (I) a cartoon of two women in fancy hats and a mirror. (J) a close up of a woman's face with a round ball on her forehead. 
\textbf{A:} (B) a man in a blue jacket and tan pants holding a sword.

\hspace{-10px}
\textbf{Q:} What is the most appropriate caption for the olmec indian? 
Answer Choices: (A) a large stone head in a garden (B) a woman lighting a candle at a table (C) [ unused0 ] is the new champion of the world (D) a table full of food (E) a group of women holding banners in a protest. (F) a woman in a traditional dress stands in front of a tent. (G) a man in a suit sitting in a chair. (H) a clay pot with a small hole on top of it. (I) a close up of three green leaves of a plant (J) a fern in the forest near the lake 
\textbf{A:} (A) a large stone head in a garden

\hspace{-10px}
\textbf{Q:} What is the most appropriate caption for the charales order? 
Answer Choices: (A) a man walking past a store with a sign on it. (B) the road to the island (C) the art of the flower (D) a black snake with white spots on it's body. (E) a bird perched on a branch in a tree. (F) a lion roaring in the wild. (G) a man falling from a skateboard (H) a large, thin, green plant with long thin leaves. (I) a close up of a bush with blue berries (J) a mouse in a hollow log. 
\textbf{A:} (H) a large, thin, green plant with long thin leaves.

\hspace{-10px}
\textbf{Q:} What is the most appropriate caption for the skink lizard? 
Answer Choices: (A) two lizards on a log with a log in the background. (B) a man walking past a store with a sign on it. (C) the art of the flower (D) a portrait of [ unused0 ], seated at a desk in front of a telescope. (E) the car is a small car that can be found in the museum. (F) a black snake with white spots on it's body. (G) a bird perched on a branch in a tree. (H) a close up of a bush with blue berries (I) a close up of a metal container with a white label on it. (J) a white door with a glass window 
\textbf{A:} (A) two lizards on a log with a log in the background.

\hspace{-10px}
\textbf{Q:} What is the most appropriate caption for the football goal? 
Answer Choices: (A) a large stadium with a large field and a large structure. (B) a woman wearing a blue ski mask (C) [ unused0 ] of spain during the fifa world cup brazil group a match between spain and argentina at the estadio santiago bernabeu on june 25, 2010 in santiago, chile. (D) a woman lifting a barbell in a competition. (E) a man in a suit and tie looking at the camera. (F) person is a player for football team (G) [ unused0 ], the brazilian national team, in the 1960s. (H) the empty stand at the old trafford stadium (I) a soccer field with a goal post in the middle (J) the emblem of the football club of the italian football club of the italian football club of the italian football club of the italian football club of the italian football club of the italian football club of the 
\textcolor{magenta}{A: (I) a soccer field with a goal post in the middle} 
 \textcolor{ForestGreen}{\ding{52}}
\\
\hline
\end{tabular}
\end{table}

\section{Conclusion}
In this work, we explore the conjunction of the Visual Word Sense Disambiguation task and Large Language Models. More specifically, our current paper is the first one to harness the rich knowledge stored in LLMs via different prompting strategies. Following the LLM-as-KB paradigm we managed to boost the performance of baseline visiolinguistic pipelines. Moreover, we examined the potential of unimodal approaches by converting VWSD to a textual question-answer problem, where generated image captions are leveraged as multiple-choice candidates. Finally, Chain-of-Thought prompting highlighted human-interpretable explainability aspects tied to the LLM-based knowledge extraction process. Overall, our analysis reveals the importance of model scale towards performing knowledge-related tasks with LLMs.

\begin{acknowledgments}
The research work was supported by the Hellenic Foundation for Research and Innovation (HFRI) under the 3rd Call for HFRI PhD Fellowships (Fellowship Number 5537). The work presented in this paper is co-funded by the European Union under the project AI4Culture, DIGITAL-2022-CULTURAL-02, Grant Agreement 101100683. 
\end{acknowledgments}

\bibliography{sample-ceur}

\appendix

\section{Scale significance for knowledge enhancement}
\label{sec:appendix1}
As an extension of Tab. \ref{tab:llm-results}, we presents results regarding LLM-enhancements using other LLMs (1.7B to 7B parameters) in Tab. \ref{tab:llm-results-appendix-1}.

Compared to Tab. \ref{tab:llm-results}, smaller models present more mediocre knowledge enhancement capabilities, frequently scoring below the knowledge-free VL retrieval baseline. Such a result indicates that knowledge enhancement is not a "free-lunch" solution to boost results, but requires tuning several aspects of the process in order to benefit the downstream task. Therefore, despite the encouraging performance of Vicuna7B/13B models, other LLMs in the lower-billion range do not possess the necessary knowledge capabilities to effectively perform knowledge enhancement for VWSD. Therefore, we can conclude that similarly to the QA prompting that demonstrated the need for scale, also prompting for knowledge enhancement is strongly associated with model size. To this end, we verify that smaller models with competitive capabilities, similar to the Vicuna series, need to be develop in order to facilitate applications related -but not limited- to LLM-based knowledge-enhancement.
\begin{table*}[h!]
\centering
\caption{Additional results for zero-shot LLM-based enhancement. }
\label{tab:llm-results-appendix-1}
\begin{tabular}{p{0.0001cm}|p{1.52cm}|p{0.45cm}p{0.55cm}|p{0.45cm}p{0.55cm}|p{0.45cm}p{0.55cm}|p{0.45cm}p{0.55cm}|p{0.45cm}p{0.55cm}|p{0.45cm}p{0.55cm}|p{0.45cm}p{0.4cm}}
\hline
\multicolumn{2}{c|}{}& \multicolumn{2}{c|}{\footnotesize \textbf{CLIP$_{LAION}$}}  
& \multicolumn{2}{c|}{\footnotesize \textbf{CLIP-L}} 
& \multicolumn{2}{c|}{\footnotesize \textbf{ALIGN}} 
&  \multicolumn{2}{c|}{\footnotesize \textbf{BLIP}$_{\footnotesize \textit{C}}$} 
& \multicolumn{2}{c|}{\footnotesize \textbf{BLIP-L}$_{\footnotesize \textit{C}}$} 
& \multicolumn{2}{c|}{\footnotesize \textbf{BLIP}$_{\footnotesize \textit{F}}$} 
& \multicolumn{2}{c}{\footnotesize \textbf{BLIP-L}$_{\footnotesize \textit{F}}$}\\
\hline
\multicolumn{2}{c|}{}& \footnotesize acc& \footnotesize MRR & \footnotesize acc& \footnotesize MRR & \footnotesize acc& \footnotesize MRR & \footnotesize acc& \footnotesize MRR & \footnotesize acc& \footnotesize MRR & \footnotesize acc& \footnotesize MRR & \footnotesize acc& \footnotesize MRR  \\
\hline

\multicolumn{16}{c}{\footnotesize \textbf{With penalty}} \\
\hline
\multicolumn{2}{c|}{\footnotesize \textbf{Baseline}} & \footnotesize 71.06 & \footnotesize 81.50 & \footnotesize 62.85 & \footnotesize 76.24  & \footnotesize 68.90 & \footnotesize 80.00 & \footnotesize 60.90 & \footnotesize 74.33 & \footnotesize 64.58 & \footnotesize 77.51 & \footnotesize 60.47 & \footnotesize 73.87 & \footnotesize 69.76 & \footnotesize 80.42
\\
\hline
\multirow{4}{0.08cm}
{\vspace{-0.13cm}\hspace{-0.15cm}\begin{turn}{90}
\notsotiny  \textbf{BLOOMZ1.7B} \end{turn}}&
\footnotesize exact &
\footnotesize  66.52 & \footnotesize  78.50 &
\footnotesize  64.92 & \footnotesize  74.50 &
\footnotesize  65.87 & \footnotesize  77.42 &
\footnotesize  64.58 & \footnotesize  76.28 &
\footnotesize  65.66 & \footnotesize  77.13 &
\footnotesize  59.18 & \footnotesize  72.70 &
\footnotesize  67.39 & \footnotesize  78.67 
\\
& \footnotesize what\_is &\footnotesize  69.98 & \footnotesize  80.67 &\footnotesize  66.74 & \footnotesize  76.41 &
\footnotesize  65.44 & \footnotesize  78.92 &
\footnotesize  63.28 & \footnotesize  75.95 &
\footnotesize  65.23 & \footnotesize  77.68 &
\footnotesize  58.32 & \footnotesize  72.06 &
\footnotesize  66.52 & \footnotesize  78.30 
\\
& \footnotesize describe &
\footnotesize  73.65 & \footnotesize  83.52 &
\footnotesize  69.84 & \footnotesize  80.08 &
\footnotesize  74.95 & \footnotesize  81.56 &
\footnotesize  66.74 & \footnotesize  78.37 &
\footnotesize  71.71 & \footnotesize  81.55 &
\footnotesize  62.63 & \footnotesize  75.55 &
\footnotesize  72.35 & \footnotesize  82.28 
\\
& \footnotesize meaning\_of & \footnotesize  69.33 & \footnotesize  80.69 &\footnotesize  65.01 & \footnotesize  76.38 &
\footnotesize  66.74 & \footnotesize  78.17 &
\footnotesize  63.50 & \footnotesize  76.44 &
\footnotesize  65.44 & \footnotesize  78.29 &
\footnotesize  58.53 & \footnotesize  72.50 &
\footnotesize  68.25 & \footnotesize  79.74 
\\
\hline

\hline
\multirow{4}{0.08cm}{\vspace{-0.13cm}\hspace{-0.15cm}\begin{turn}{90}
\notsotiny \textbf{OPT2.7B} \end{turn}}& \footnotesize exact &
\footnotesize  71.06 & \footnotesize  81.46 &\footnotesize  62.85 & \footnotesize  76.00 &
\footnotesize  68.68 & \footnotesize  75.93 &
\footnotesize  61.12 & \footnotesize  74.46 &
\footnotesize  64.58 & \footnotesize  77.41 &
\footnotesize  60.26 & \footnotesize  73.73 &
\footnotesize  69.76 & \footnotesize  80.36 
\\
& \footnotesize what\_is &
\footnotesize  66.95 & \footnotesize  78.99 &\footnotesize  66.30 & \footnotesize  74.85 &
\footnotesize  63.28 & \footnotesize  78.10 &
\footnotesize  60.91 & \footnotesize  74.43 &
\footnotesize  66.31 & \footnotesize  77.86 &
\footnotesize  57.24 & \footnotesize  71.15 &
\footnotesize  67.60 & \footnotesize  78.58 
\\
& \footnotesize describe &
\footnotesize  68.03 & \footnotesize  79.89 &\footnotesize  66.08 & \footnotesize  74.75 &
\footnotesize  64.79 & \footnotesize  78.14 &
\footnotesize  61.77 & \footnotesize  74.73 &
\footnotesize  66.31 & \footnotesize  77.57 &
\footnotesize  57.67 & \footnotesize  71.48 &
\footnotesize  68.03 & \footnotesize  79.03 
\\
& \footnotesize meaning\_of &
\footnotesize  69.11 & \footnotesize  80.56 &\footnotesize  65.25 & \footnotesize  75.60 &
\footnotesize  65.66 & \footnotesize  77.45 &
\footnotesize  61.99 & \footnotesize  75.35 &
\footnotesize  63.93 & \footnotesize  76.88 &
\footnotesize  58.32 & \footnotesize  71.65 &
\footnotesize  65.44 & \footnotesize  77.69 \\
\hline

\multirow{4}{0.08cm}{\vspace{-0.13cm}\hspace{-0.15cm}\begin{turn}{90}
\notsotiny \textbf{BLOOMZ3B} \end{turn}}& \footnotesize exact &
\footnotesize  67.82 & \footnotesize  79.31 &\footnotesize  62.99 & \footnotesize  74.59 &
\footnotesize  66.52 & \footnotesize  76.18 &
\footnotesize  60.48 & \footnotesize  73.13 &
\footnotesize  63.28 & \footnotesize  76.00 &
\footnotesize  57.02 & \footnotesize  71.23 &
\footnotesize  65.66 & \footnotesize  77.49 
\\
& \footnotesize what\_is &
\footnotesize  71.92 & \footnotesize  81.78 &\footnotesize  68.25 & \footnotesize  76.82 &
\footnotesize  67.39 & \footnotesize  79.82 &
\footnotesize  61.34 & \footnotesize  74.94 &
\footnotesize  66.95 & \footnotesize  78.47 &
\footnotesize  59.61 & \footnotesize  73.35 &
\footnotesize  68.47 & \footnotesize  79.58 
\\
& \footnotesize describe &
\footnotesize  70.84 & \footnotesize  81.11 &\footnotesize  65.28 & \footnotesize  75.38 &
\footnotesize  66.09 & \footnotesize  78.07 &
\footnotesize  62.85 & \footnotesize  75.65 &
\footnotesize  67.39 & \footnotesize  78.71 &
\footnotesize  57.24 & \footnotesize  71.72 &
\footnotesize  67.82 & \footnotesize  79.20 
\\
& \footnotesize meaning\_of &
\footnotesize  70.63 & \footnotesize  81.47 &\footnotesize  67.32 & \footnotesize  77.96 &
\footnotesize  68.47 & \footnotesize  78.76 &
\footnotesize  63.71 & \footnotesize  76.52 &
\footnotesize  66.31 & \footnotesize  78.55 &
\footnotesize  59.40 & \footnotesize  73.60 &
\footnotesize  68.03 & \footnotesize  79.26 \\
\hline

\multirow{4}{0.08cm}{\vspace{-0.13cm}\hspace{-0.15cm}\begin{turn}{90}
\notsotiny \textbf{OPT6.7B} \end{turn}}&
\footnotesize exact & \footnotesize  62.63 & \footnotesize  75.84 &
\footnotesize  62.20 & \footnotesize  75.54 &
\footnotesize  67.82 & \footnotesize  79.24 &
\footnotesize  60.91 & \footnotesize  74.23 &
\footnotesize  64.79 & \footnotesize  77.58 &
\footnotesize  59.83 & \footnotesize  73.40 &
\footnotesize  69.11 & \footnotesize  79.94
\\
& \footnotesize what\_is & \footnotesize  61.79 & \footnotesize  75.70 &
\footnotesize  64.63 & \footnotesize  77.68 &
\footnotesize  64.79 & \footnotesize  77.23 &
\footnotesize  61.77 & \footnotesize  75.01 &
\footnotesize  63.07 & \footnotesize  76.16 &
\footnotesize  57.88 & \footnotesize  71.79 &
\footnotesize  65.87 & \footnotesize  77.77 
\\
& \footnotesize describe & \footnotesize  64.43 & \footnotesize  76.91 &
\footnotesize  65.73 & \footnotesize  78.24 &
\footnotesize  65.23 & \footnotesize  77.89 &
\footnotesize  61.12 & \footnotesize  74.67 &
\footnotesize  63.93 & \footnotesize  77.07 &
\footnotesize  56.16 & \footnotesize  71.30 &
\footnotesize  66.09 & \footnotesize  78.38 
\\
& \footnotesize meaning\_of & \footnotesize  62.17 & \footnotesize  75.84 &
\footnotesize  63.61 & \footnotesize  77.19 &
\footnotesize  66.74 & \footnotesize  78.47 &
\footnotesize  63.28 & \footnotesize  75.93 &
\footnotesize  65.44 & \footnotesize  77.43 &
\footnotesize  59.83 & \footnotesize  72.96 &
\footnotesize  68.03 & \footnotesize  78.75 
\\
\hline

\multirow{4}{0.08cm}{\vspace{-0.13cm}\hspace{-0.15cm}\begin{turn}{90}
\notsotiny \textbf{LLAMA7B} \end{turn}}& \footnotesize exact &
\footnotesize  62.20 & \footnotesize  76.04 &\footnotesize  57.45 & \footnotesize  70.70 &
\footnotesize  60.48 & \footnotesize  72.23 &
\footnotesize  52.92 & \footnotesize  68.76 &
\footnotesize  54.00 & \footnotesize  69.51 &
\footnotesize  48.81 & \footnotesize  65.54 &
\footnotesize  58.10 & \footnotesize  72.33 
\\
& \footnotesize what\_is  &
\footnotesize  68.90 & \footnotesize  80.48 &
\footnotesize  67.10 & \footnotesize  77.26 &
\footnotesize  66.95 & \footnotesize  78.93 &
\footnotesize  68.03 & \footnotesize  68.03 &
\footnotesize  66.95 & \footnotesize  78.48 &
\footnotesize  58.75 & \footnotesize  73.29 &
\footnotesize  70.84 & \footnotesize  80.84
\\
& \footnotesize describe &
\footnotesize  72.14 & \footnotesize  82.02 &\footnotesize  66.81 & \footnotesize  76.80 &
\footnotesize  69.76 & \footnotesize  79.04 &
\footnotesize  64.79 & \footnotesize  76.58 &
\footnotesize  66.09 & \footnotesize  77.99 &
\footnotesize  58.53 & \footnotesize  72.44 &
\footnotesize  69.11 & \footnotesize  80.15 
\\
& \footnotesize meaning\_of &
\footnotesize  67.82 & \footnotesize  80.11 &
\footnotesize  62.90 & \footnotesize  76.12 &
\footnotesize  65.66 & \footnotesize  76.53 &
\footnotesize  60.26 & \footnotesize  74.40 &
\footnotesize  62.42 & \footnotesize  75.71 &
\footnotesize  54.00 & \footnotesize  69.84 &
\footnotesize  65.87 & \footnotesize  78.19 
\\
\hline

\multicolumn{16}{c}{\footnotesize \textbf{Without penalty}} \\
\hline
\multicolumn{2}{c|}{\footnotesize \textbf{Baseline}} & \footnotesize 67.82 & \footnotesize 79.50 & \footnotesize 60.69 & \footnotesize 74.42 & \footnotesize 65.66 & \footnotesize 77.48 & \footnotesize 57.24 & \footnotesize 72.07 & \footnotesize 61.34 & \footnotesize 75.88 & \footnotesize 57.67 & \footnotesize 71.96 & \footnotesize 65.01 & \footnotesize 77.86
\\
\hline
\multirow{4}{0.08cm}{\vspace{-0.13cm}\hspace{-0.15cm}\begin{turn}{90}
\notsotiny \textbf{BLOOMZ 1.7B} \end{turn}}& \footnotesize exact &
\footnotesize  63.71 & \footnotesize  76.53 &
\footnotesize  61.66 & \footnotesize  72.23 &
\footnotesize  63.28 & \footnotesize  75.05 &
\footnotesize  59.18 & \footnotesize  72.96 &
\footnotesize  62.85 & \footnotesize  74.99 &
\footnotesize  56.37 & \footnotesize  70.60 &
\footnotesize  63.50 & \footnotesize  76.20 
\\
& \footnotesize what\_is & \footnotesize  67.60 & \footnotesize  79.04 & \footnotesize  65.01 & \footnotesize  75.30 &
\footnotesize  63.07 & \footnotesize  77.33 &
\footnotesize  59.40 & \footnotesize  73.50 &
\footnotesize  62.85 & \footnotesize  76.00 &
\footnotesize  56.16 & \footnotesize  70.54 &
\footnotesize  65.23 & \footnotesize  77.26 
\\
& \footnotesize describe & 
\footnotesize  67.60 & \footnotesize  79.54 &
\footnotesize  62.99 & \footnotesize  74.68 &
\footnotesize  66.52 & \footnotesize  76.05 &
\footnotesize  59.83 & \footnotesize  74.59 &
\footnotesize  62.63 & \footnotesize  76.47 &
\footnotesize  53.56 & \footnotesize  69.84 &
\footnotesize  63.71 & \footnotesize  77.06 
\\
& \footnotesize meaning\_of & \footnotesize  66.52 & \footnotesize  78.70 & \footnotesize  64.15 & \footnotesize  73.78 &
\footnotesize  64.36 & \footnotesize  77.03 &
\footnotesize  60.48 & \footnotesize  74.34 &
\footnotesize  61.99 & \footnotesize  76.13 &
\footnotesize  56.16 & \footnotesize  70.61 &
\footnotesize  65.01 & \footnotesize  77.89 
\\
\hline

\hline
\multirow{4}{0.08cm}{\vspace{-0.13cm}\hspace{-0.15cm}\begin{turn}{90}
\notsotiny \textbf{OPT2.7B} \end{turn}}& \footnotesize exact &
\footnotesize  67.82 & \footnotesize  79.47 &\footnotesize  60.26 & \footnotesize  72.77 &
\footnotesize  65.66 & \footnotesize  74.15 &
\footnotesize  57.45 & \footnotesize  72.19 &
\footnotesize  61.12 & \footnotesize  75.77 &
\footnotesize  57.24 & \footnotesize  71.68 &
\footnotesize  65.01 & \footnotesize  77.90 
\\
& \footnotesize what\_is &
\footnotesize  65.44 & \footnotesize  77.60 &\footnotesize  62.75 & \footnotesize  72.91 &
\footnotesize  61.12 & \footnotesize  75.47 &
\footnotesize  59.83 & \footnotesize  73.13 &
\footnotesize  61.12 & \footnotesize  74.54 &
\footnotesize  53.35 & \footnotesize  68.71 &
\footnotesize  63.50 & \footnotesize  76.22 
\\
& \footnotesize describe &
\footnotesize  65.87 & \footnotesize  78.09 &\footnotesize  63.89 & \footnotesize  72.95 &
\footnotesize  62.20 & \footnotesize  76.31 &
\footnotesize  59.83 & \footnotesize  73.28 &
\footnotesize  62.20 & \footnotesize  75.17 &
\footnotesize  54.43 & \footnotesize  69.86 &
\footnotesize  63.28 & \footnotesize  76.28 
\\
& \footnotesize meaning\_of &
\footnotesize  67.60 & \footnotesize  79.28 &\footnotesize  62.99 & \footnotesize  72.97 &
\footnotesize  64.58 & \footnotesize  75.79 &
\footnotesize  59.18 & \footnotesize  73.38 &
\footnotesize  60.26 & \footnotesize  74.70 &
\footnotesize  54.86 & \footnotesize  69.43 &
\footnotesize  62.42 & \footnotesize  75.86 \\
\hline

\multirow{4}{0.08cm}{\vspace{-0.13cm}\hspace{-0.15cm}\begin{turn}{90}
\notsotiny \textbf{BLOOMZ3B} \end{turn}}& \footnotesize exact &
\footnotesize  64.15 & \footnotesize  76.98 &\footnotesize  59.52 & \footnotesize  71.53 &
\footnotesize  63.93 & \footnotesize  73.78 &
\footnotesize  58.10 & \footnotesize  71.77 &
\footnotesize  59.61 & \footnotesize  74.06 &
\footnotesize  54.86 & \footnotesize  69.66 &
\footnotesize  61.12 & \footnotesize  74.99 
\\
& \footnotesize what\_is &
\footnotesize  69.11 & \footnotesize  80.03 &\footnotesize  65.66 & \footnotesize  75.39 &
\footnotesize  62.85 & \footnotesize  77.88 &
\footnotesize  61.34 & \footnotesize  74.35 &
\footnotesize  65.01 & \footnotesize  77.32 &
\footnotesize  57.24 & \footnotesize  71.85 &
\footnotesize  68.03 & \footnotesize  79.12 
\\
& \footnotesize describe &
\footnotesize  68.68 & \footnotesize  79.47 &\footnotesize  62.88 & \footnotesize  73.83 &
\footnotesize  63.50 & \footnotesize  76.11 &
\footnotesize  60.48 & \footnotesize  73.87 &
\footnotesize  62.85 & \footnotesize  76.06 &
\footnotesize  54.86 & \footnotesize  70.48 &
\footnotesize  65.66 & \footnotesize  77.64 
\\
& \footnotesize meaning\_of &
\footnotesize  68.25 & \footnotesize  79.73 &\footnotesize  64.94 & \footnotesize  75.51 &
\footnotesize  66.31 & \footnotesize  77.17 &
\footnotesize  61.77 & \footnotesize  74.92 &
\footnotesize  62.42 & \footnotesize  76.27 &
\footnotesize  57.02 & \footnotesize  71.79 &
\footnotesize  65.23 & \footnotesize  77.21 \\
\hline

\multirow{4}{0.08cm}{\vspace{-0.13cm}\hspace{-0.15cm}\begin{turn}{90}
\notsotiny \textbf{OPT 6.7B} \end{turn}}&
\footnotesize exact & \footnotesize  58.75 & \footnotesize  72.63 &
\footnotesize  59.61 & \footnotesize  73.86 &
\footnotesize  64.15 & \footnotesize  76.57 &
\footnotesize  57.24 & \footnotesize  71.96 &
\footnotesize  61.12 & \footnotesize  75.83 &
\footnotesize  56.80 & \footnotesize  71.40 &
\footnotesize  64.79 & \footnotesize 77.66
\\
& \footnotesize what\_is & \footnotesize  60.48 & \footnotesize  74.10 &
\footnotesize  62.45 & \footnotesize  75.89 &
\footnotesize  61.77 & \footnotesize  75.18 &
\footnotesize  57.88 & \footnotesize  72.27 &
\footnotesize  61.77 & \footnotesize  74.89 &
\footnotesize  52.92 & \footnotesize  68.83 &
\footnotesize  61.99 & \footnotesize  75.23 
\\
& \footnotesize describe & \footnotesize  60.74 & \footnotesize  74.28 &
\footnotesize  63.12 & \footnotesize  76.19 &
\footnotesize  63.28 & \footnotesize  76.26 &
\footnotesize  59.40 & \footnotesize  73.03 &
\footnotesize  58.96 & \footnotesize  73.86 &
\footnotesize  52.92 & \footnotesize  69.18 &
\footnotesize  62.63 & \footnotesize  76.13 
\\
& \footnotesize meaning\_of & \footnotesize  59.28 & \footnotesize  73.77 &
\footnotesize  62.17 & \footnotesize  76.04 &
\footnotesize  63.71 & \footnotesize  76.31 &
\footnotesize  52.92 & \footnotesize  74.37 &
\footnotesize  61.99 & \footnotesize  75.47 &
\footnotesize  55.94 & \footnotesize  70.67 &
\footnotesize  65.01 & \footnotesize  77.27 
\\
\hline

\multirow{4}{0.08cm}{\vspace{-0.13cm}\hspace{-0.15cm}\begin{turn}{90}
\notsotiny \textbf{LLAMA7B} \end{turn}}& \footnotesize exact & \footnotesize  60.91 & \footnotesize  74.66 &\footnotesize  56.16 & \footnotesize  68.98 &
\footnotesize  56.80 & \footnotesize  70.76 &
\footnotesize  52.05 & \footnotesize  67.33 &
\footnotesize  50.54 & \footnotesize  66.81 &
\footnotesize  47.95 & \footnotesize  64.17 &
\footnotesize  54.86 & \footnotesize  69.79 
\\
& \footnotesize what\_is &
\footnotesize  67.60 & \footnotesize  79.19 &\footnotesize  64.71 & \footnotesize  75.98 &
\footnotesize  66.31 & \footnotesize  77.31 &
\footnotesize  64.36 & \footnotesize  76.30 &
\footnotesize  64.36 & \footnotesize  77.00 &
\footnotesize  56.37 & \footnotesize  71.90 &
\footnotesize  68.47 & \footnotesize  79.43 
\\
& \footnotesize describe &
\footnotesize  70.41 & \footnotesize  81.20 &\footnotesize  63.77 & \footnotesize  74.97 &
\footnotesize  66.74 & \footnotesize  76.99 &
\footnotesize  64.15 & \footnotesize  76.25 &
\footnotesize  63.50 & \footnotesize  76.40 &
\footnotesize  58.10 & \footnotesize  72.58 &
\footnotesize  68.25 & \footnotesize  79.36 
\\
& \footnotesize meaning\_of &
\footnotesize  66.09 & \footnotesize  78.60 &\footnotesize  60.41 & \footnotesize  72.84 &
\footnotesize  62.85 & \footnotesize  74.79 &
\footnotesize  58.75 & \footnotesize  72.70 &
\footnotesize  58.53 & \footnotesize  73.12 &
\footnotesize  52.70 & \footnotesize  68.03 &
\footnotesize  62.42 & \footnotesize  76.07 
\\
\hline

\end{tabular}

\end{table*}

\section{Reproducibility}
Our code can be found in GitHub: \href{https://github.com/anastasiakrith/llm-for-vwsd/}{https://github.com/anastasiakrith/llm-for-vwsd/}.

\end{document}